# ARC: Autonomous Robotics Compliance

A Three-Layer Governance Architecture for Deployed Autonomous Systems

Tord Eide (CRAB) · Einar Holt (LEA) · Wybe Labs Research Team

Wybe Labs Inc. · Moss, Norway / USA · September 9, 2026

*PREPRINT — NOT YET PEER REVIEWED*

## Abstract

The commercial deployment of autonomous robotic systems in safety-critical environments has outpaced the development of adequate governance frameworks. A fundamental conflation underlies this gap: the assumption that demonstrated capability constitutes sufficient authorization for deployment. We argue that governance requires three distinct but interdependent layers: (1) model-level safety validation — does the AI model respond appropriately to physical safety constraints? (2) system-level cognitive certification — does the integrated system meet domain-specific capability thresholds, evaluated independently? (3) operational-level authorization — is this specific robot permitted to perform this specific task at this specific site?

We propose **ARC — Autonomous Robotics Compliance**, instantiating these layers through ASIMOV-Agentic (Google DeepMind, 2025/2026), CRAB — Cognitive Robotics Accreditation Benchmarks (Wybe Labs, 2026), and LEA-1:2026 — Levels of Earned Autonomy (Wybe Labs, 2026). Empirical evidence from ASIMOV-2.0 demonstrates that frontier AI models exhibit constraint violation rates of 30% or greater under embodiment-specific safety conditions, motivating the upper two layers. ARC is a testable governance proposal — not an established certification or authorization system — intended to trigger peer review, empirical testing, and regulatory engagement.

## 1. Introduction

In 2026, autonomous robotic systems have crossed from controlled demonstrations into operational deployment at scale. Agility Robotics reports commercial deployments of Digit in logistics facilities and has announced a proposed public listing [36]. Diligent Robotics reports more than 1.25 million autonomous hospital deliveries with its Moxi platform [37]. 1X Technologies is shipping its bipedal NEO home platform [38]. Google DeepMind and Apptronik have demonstrated whole-body humanoid control capable of multi-step task execution in unstructured environments [1]. The industry has reached the threshold at which independent governance is no longer a research aspiration — it is an operational necessity.

The frameworks available to deployers, regulators, and investors remain inadequate. Manufacturers publish curated demonstrations. Investors rely on narrative. Hospitals deploy systems whose operational boundaries they cannot independently verify. At the centre of this governance gap lies a conceptual conflation: **the equation of capability with permission**. When a manufacturer demonstrates that a system can navigate a corridor, deliver a meal tray, or assist with patient repositioning, this is treated as establishing that the system *should* perform that task in a given deployment context. The distinction between what a system *can* do and what it is *authorized* to do — under specific conditions, for a specific population, in a specific regulatory environment — has not been operationally formalized in any existing framework.

This paper proposes **ARC — Autonomous Robotics Compliance**: a three-layer governance architecture that closes this gap. ARC is named deliberately: the arc spans the full governance distance from model benchmark to bedside deployment, connecting safety science, independent certification, and site-specific authorization into a coherent whole. The three layers are:

- **Layer 1 — Model Validation (ASIMOV):** Does the underlying AI model respond safely to physical safety constraints and uncertainty?
- **Layer 2 — System Certification (CRAB):** Does the integrated system meet domain-specific cognitive capability thresholds, evaluated independently?
- **Layer 3 — Operational Authorization (LEA):** Is this specific robot authorized to perform this specific task at this specific site, and on what evidentiary basis?

## 2. Background and Related Work

### 2.1 What Exists — and What It Does Not Cover

**ISO TC 299** has produced substantive standards covering physical safety: ISO 10218-1:2025 and ISO 10218-2:2025 for industrial robots, ISO 13482:2014 for personal care robots, and the developing ISO/CD 25785-1 for dynamically stable mobile robots [10,11,33,34]. These evaluate what robots do physically — collision forces, emergency stop protocols, envelope constraints. They do not address cognitive capability: the capacity of a system to recognize the limits of its authority, handle ambiguous instructions, or reason about when it should decline to act.

**IEEE** is the most relevant active standards body for cognitive dimensions. IEEE P3777 addresses benchmarking and performance metrics for AI agents; IEEE P2940 covers robot agility; IEEE P3704 addresses evaluation methods for embodied-intelligence capabilities [12,13,35]. This work is methodologically serious. None of these projects currently produces certifiable cognitive capability assessment for real-world deployed systems with cryptographically verifiable attestation, nor do they address the operational authorization question.

**Academic benchmarks** — LIBERO, RoboCasa, ManiSkill2, SIMPLER, GOAT-Bench, EmbodiedBench, ManipBench — provide rigorous methods for controlled research comparison [2–4, 29–32]. They do not produce transferable certificates that a hospital procurement team or regulatory body can rely on.

A qualitative analysis of 20 recent European research projects on elder care robotics found that ethical and legal dimensions receive systematically limited attention, and that robots are frequently treated as external products rather than actors within accountability structures [40].

Human-robot interaction taxonomies [14–19], healthcare acceptance studies [20–22], runtime-assurance research [23–25], and the NIST AI RMF [26] all provide relevant conceptual foundations that ARC draws on and connects into an operational governance architecture.

### 2.2 The ASIMOV Evidence Base

The most directly relevant empirical contribution is ASIMOV-2.0, published by Jindal, Kalashnikov et al. at Google DeepMind Robotics [5]. The paper asks: how well do frontier AI models understand physical safety constraints when operating as embodied agents?

Three benchmark components:

- **ASIMOV-Injury:** Do models understand physical risk and injury severity from text or image/video descriptions?
- **ASIMOV-Constraints:** Do models adhere to embodiment-specific safety instructions — force limits, reach constraints, operating parameters for a particular robot body?
- **ASIMOV-Video:** Do models perceive evolving physical risk in real-time video?

The central finding: no major frontier model — including GPT-5, Gemini-2.5-Pro, and Claude Opus 4.1 — achieves a constraint violation rate below 30% when reasoning simultaneously about embodiment-specific limitations, physics, and visual inputs. Three structural gaps are identified: a **modality gap** (text comprehension exceeds video/action performance), an **embodiment gap** (no model reasons adequately about a specific robot body's constraints), and a **latency gap** (smaller on-device models score lower than large cloud models).

These findings were incorporated into the Gemini Robotics 2 release of July 30, 2026 [6], which introduced ASIMOV-Agentic as a first-class evaluation component.

ASIMOV-Agentic is a necessary but insufficient governance layer. It validates AI model behaviour under test conditions but does not produce system-level certification or address what a specific robot is authorized to do at a specific site. These are the gaps CRAB and LEA address. Together — ASIMOV + CRAB + LEA — they form the complete ARC.

## 3. The ARC Architecture

### 3.1 Overview

| Layer | Governance Question | Output | Evaluated by |
| --- | --- | --- | --- |
| 1 — ASIMOV Model Validation | Does the AI model behave safely under physical safety constraints? | Benchmark scores; constraint violation rate per scenario class | AI laboratory or benchmark consortium |
| 2 — CRAB System Certification | Does the integrated system meet domain-specific cognitive capability thresholds, evaluated independently? | Per-dimension capability profile; domain-specific assessment outcome | Independent accredited review body |
| 3 — LEA Operational Authorization | Is this specific robot authorized to perform this specific task class at this specific site, on what evidentiary basis? | Authorization record binding one robot, one task class, one site — with provisional or earned status | Site operator and named responsible person, subject to applicable oversight model |

The layers are interdependent: Layer 1 outputs constrain Layer 2 evaluation design; Layer 2 profiles constrain Layer 3 authorization ceilings. No single layer, and no pair excluding Layer 3, answers the operational authorization question. No pair excluding Layer 2 answers the system certification question. Full governance — the complete ARC — requires all three.

## 3.2 Layer 1 — Model Validation (ASIMOV)

Before deployment consideration, the AI model powering the platform must be evaluated on ASIMOV-Agentic against scenarios calibrated to the intended contact class and environment — not a generic benchmark result, but an evaluation of the specific model version to be deployed against the embodiment-specific constraints of that deployment context.

Layer 1 carries structural limitations that motivate the upper two layers. Benchmark performance is conditional on the test distribution: a model may score well on ASIMOV scenarios while failing on out-of-distribution physical configurations. ASIMOV evaluates model-level behaviour — sensor fusion errors, actuation noise, and compounding real-world disturbances are not captured. ASIMOV does not address operational continuity over multi-hour shifts. Layer 1 outputs are necessary conditions for proceeding to Layer 2. They are not sufficient conditions for deployment authorization.

## 3.3 Layer 2 — System Certification (CRAB)

CRAB — Cognitive Robotics Accreditation Benchmarks [7] — proposes the first independent cognitive certification framework for deployed autonomous systems. Like aircraft certification — not performed by the manufacturer, not reduced to a single score, not valid indefinitely — CRAB evaluates the integrated system, not the model in isolation. CRAB is a draft proposal; its claims, thresholds, and governance arrangements remain open to peer review and revision.

### 3.3.1 Eight Cognitive Dimensions

**Tier 1 — Core Cognitive Capabilities:**

| Code | Dimension | What It Measures |
|---|---|---|
| SA | Situational Awareness | Environmental understanding, object and context recognition, scene inference under ambiguity |

| UP | Planning & Execution | Multi-step goal decomposition, plan revision under changing conditions, execution recovery |
|---|---|---|
| HI | Human Interaction | Natural language understanding, intent recognition, appropriate social and emotional response |
| SC | Self-Correction | Error detection, fault recovery, adaptive re-planning without human intervention |
| CG | Generalization | Performance on tasks and environments outside the training distribution |
| CE | Energy & Efficiency | Resource optimization, runtime endurance, autonomous recharge and task resumption |

**Tier 2 — Trust and Governance Capabilities (double-weighted in CRAB-H; mandatory minimum thresholds):**

| Code | Dimension | What It Measures |
|---|---|---|
| SCA ^H | Safety & Constraint Adherence | Refusal of requests outside the system's authorized scope; recognition of authority limits and role boundaries; escalation protocols; uncertainty handling; regulatory and policy compliance. Central question: can the system recognize when it should not act? |
| DT ^H | Decision Transparency | Post-action reasoning summaries; traceability of decision factors to observable inputs; consistency between stated reasoning and actual behaviour; auditability by non-technical supervisors |

**SCA in practice:** A care home resident says "Give me my insulin." The correct system response is not compliance. It is recognition that medication administration falls outside the system's authorized scope, graceful refusal, escalation to nursing staff, and documentation of the interaction — all without distressing the resident. This requires modelling authority structures, understanding regulatory constraints, reasoning about operational limits, and exercising judgment under uncertainty. ASIMOV demonstrates that frontier models fail this at the model level with >30% violation rates. CRAB-H certification tests whether the integrated system achieves it consistently across the full healthcare scenario space. CRAB scenario design for SCA draws on Murashova et al. (2025), who demonstrate that effective ethical risk assessment in Norwegian care settings requires iterative, cross-disciplinary workshops with documented dissent rather than single-pass checklists [41].

**DT in practice:** After acting, a CRAB-certified system should be able to produce: "I chose route B because route A had a stationary obstacle and a patient in apparent distress was detected at waypoint 3. I notified the nursing station at 02:14:37." That is transparency. Van Otterdijk et al. (2025) demonstrate that movement, form, and behavioural expression jointly determine how humans interpret robot intent — and that misinterpretation is common even in brief interactions [44]. CRAB-H DT scenarios test not only post-hoc reasoning summaries, but whether real-time behavioural signals are correctly understood by the resident population.

#### 3.3.2 Domain Suites

- **CRAB-H (Healthcare):** For hospitals, care homes, rehabilitation centres, and assisted living facilities. Certification scenarios include medication and supply delivery (authorized versus unauthorized request handling), fall detection and escalation, patient interaction in cognitive decline and distress contexts, privacy-sensitive information handling, emergency response coordination, and night shift autonomous operation across 8–12 hour evaluation windows. SCA and DT are double-weighted and carry mandatory minimum thresholds. A system cannot achieve CRAB-H without meeting the SCA floor. For CRAB-H, accredited review bodies may additionally publish domain-specific minimum floors for Tier 1 dimensions. Any such floor must be published in advance of evaluation and applied consistently. This case study applies an illustrative CG floor of 750 for the meal-round-logistics task class in an E1 environment.
- **CRAB-A (Administrative and Coordination):** For autonomous systems handling scheduling, coordination, communication, and administrative workflows. Scenarios include calendar management under competing constraints, policy-compliant communication drafting, PII recognition and protection, multi-agent workflow coordination, and escalation to human decision-makers at appropriate thresholds.
- **CRAB-I (Industrial and Logistics):** For warehouse, manufacturing, assembly, and logistics deployments. Scenarios include order picking under dynamic inventory changes, assembly task generalization to new component variants, quality inspection and anomaly escalation, human-robot collaborative workflows, and safety zone enforcement in multi-agent environments.

#### 3.3.3 Capability Profile and Attestation

Every CRAB evaluation produces a per-dimension capability profile as the primary governance evidence. Any aggregate summary derived from dimension scores is subordinate to the profile and does not constitute a certification basis; authorization decisions must be grounded in the per-dimension record. Cryptographic attestation via the Cardano blockchain is proposed as a tamper-evident anchoring mechanism for signed assessment records. A transaction hash provides evidence that a credential existed at a specific time and has not been altered; it does not independently validate evaluator competence, test validity, or continuing status. Full verifiability requires an agreed W3C-compatible architecture: decentralized identifiers [27], verifiable credentials [28], trusted issuer registries, revocation services, and defined audit access rules.

### 3.4 Layer 3 — Operational Authorization (LEA)

LEA-1:2026 — Levels of Earned Autonomy [8] — addresses the operational authorization gap. Its fundamental design decision: **LEA classifies grants, not robots.** A grant is an authorization that binds together three elements: one identified robot, one task class, and one site. The same physical platform may hold multiple simultaneous grants with different authorization levels for different task classes and environments. A robot authorized for autonomous logistics in a controlled warehouse holds a different grant — at a different authorization level — than it would at a residential care facility, even if its CRAB profile is identical in both contexts. Capability is invariant; authorization is context-dependent.

#### 3.4.1 Five Grant Dimensions

**Autonomy Level (A0–A5):** From teleoperated (A0) through assisted execution (A1), supervised task autonomy (A2), conditional autonomy (A3), limited high autonomy (A4), to open autonomy (A5). A5 carries no defined conformance path — full autonomous operation without any supervision or envelope constraint is not yet supportable by available evidence. The absence of a conformance path is a governance feature, not an oversight.

**Contact Class (C0–C3):** The most intimate form of human contact the task class requires. C0 (no contact), C1 (proximity and interaction), C2 (physical assistance — capped at A2 with mandatory live supervision), C3 (clinical and corporeal actions — capped at A1; clinical actions remain the actions of authorized healthcare personnel, with the robot at most an instrument under continuous supervision). The C2 and C3 caps are grounded in Saplacan and Tørresen (2022): physical assistance tasks require situational context that autonomous systems cannot reliably provide without live human oversight [42].

**Environment Class (E0–E2):** Controlled industrial (E0), certified institution (E1), private household (E2). Evidence accumulated at a lower environment class does not transfer to a higher-numbered class. A grant earned at E1 does not support a corresponding grant at E2; household deployment cannot be justified by institutional evidence [42].

**Learning Mode (LM0–LM3):** Frozen (LM0, maximum A4), gated updates (LM1, maximum A4), on-site learning (LM2, maximum A3), fleet learning (LM3, maximum A2 unless per-site gated). The learning mode axis addresses a problem that point-in-time certification cannot: an AI-based worker is designed to change. A system operating in LM2 mode is genuinely different after one month of deployment than at initial assessment. Learning mode caps impose a structural constraint that limits the autonomy level available to systems whose behaviour may drift between explicit revalidation events.

**Status:** PROVISIONAL → EARNED → SUSPENDED → REVOKED. Earned status is not declared — it is acquired through logged supervised operation meeting specified thresholds, and maintained through ongoing metric monitoring. A grant that fails its maintenance conditions reverts to SUSPENDED and drops one autonomy level automatically, without requiring a human decision. This continuous validity condition addresses the static certification problem directly: there is no certification moment at which a grant is valid indefinitely.

#### 3.4.2 Fallback Ladder and Incident Response

The **fallback ladder (F1–F5)** mandates a structured response to any exception: securing held objects (F1), announcing status (F2), withdrawing to a safe position (F3), notifying the responsible person (F4), waiting for human intervention (F5). The ladder is obligatory — resolution by chance rather than following the ladder counts against fallback integrity metrics and can trigger status changes.

**Incident severity (S1/S2/S3)** determines automatic responses: S1 (procedural deviation, no injury) is logged and counted against fallback integrity; S2 (minor injury or near-miss) triggers automatic suspension; S3 (injury requiring treatment, or any harm to a vulnerable person) triggers automatic revocation and mandatory reporting.

#### 3.4.3 Graduation Thresholds

Graduation from PROVISIONAL to EARNED at the illustrative A3/C1/E1 level requires, over a continuous assessment window:

- ≥ 400 supervised operational hours in the task class at the site
- Intervention rate ≤ 0.25 per 100 hours (attributable to robot behaviour)
- Fallback integrity ≥ 98%
- No unresolved incident of S2 or higher severity

These values are informed governance proposals — proportionate to the risk of conditional autonomous operation in an institutional environment with vulnerable occupants — not empirically validated optima. They should be revised as deployment evidence becomes available.

## 4. Capability Is Not Permission

### 4.1 The Core Distinction

Capability is a property of the system — what it can do given its current state, training, and hardware configuration. Permission is a relational property — what it is authorized to do given its deployment context, the regulatory environment, the characteristics of the population it serves, and the oversight arrangements in place.

The conflation of capability with permission is a category error with practical consequences. A system capable of administering medication is not thereby authorized to do so in a clinical environment. A system capable of autonomous warehouse operation is not thereby authorized to operate at the same autonomy level in a care home. The capability is invariant; the authorization is context-dependent.

### 4.2 Why the Distinction Becomes More Critical as Capability Increases

A more capable system can cause more harm without appropriate authorization. A system at A0 (teleoperated) can cause harm only when its human operator does. A system at A3 (conditional autonomy) can cause harm through its own judgment errors over extended periods without observation. A system at A4 (limited high autonomy) can cause harm at scale before intervention is possible. The governance burden — the evidence required to justify a given authorization level — should therefore increase monotonically with the autonomy level being sought. LEA operationalizes this: higher autonomy levels require more supervised hours, lower intervention rates, and higher fallback integrity before graduation to EARNED status.

### 4.3 The Static Certification Problem

AI-based workers are designed to change — through on-site learning, remote updates, fleet learning, and environmental adaptation. A system assessed at t=0 may be substantively different at t=90 days in LM2 mode. Existing certification frameworks treat assessment as a point-in-time event, adequate for physical products that do not change between production and use. LEA's earned-status system addresses this architecturally: authorization validity depends on current operational metrics, not past certification moments. The authorization expires de facto when the performance evidence no longer supports it.

## 5. Healthcare Deployment Case Study

### 5.1 Scenario and Sensor Selection

A mobile service-robot platform — illustratively, a wheeled platform such as Robot.com R-noid, or a separately evaluated bipedal platform such as 1X NEO — is proposed for deployment in a certified residential care facility (E1 environment) for meal delivery and logistics. The platform operates with LM1 learning mode (gated updates only; behaviour changes only through versioned releases with regression testing).

Primary sensing uses mono-thermal and depth sensors; RGB video is restricted to time-limited fault diagnosis under explicit authorization. Baselizadeh et al. (2024) demonstrate that mono-thermal cameras achieve the best balance between technical utility and privacy perception in care robot sensing, while RGB sensors are experienced as more intrusive by both younger and older residents [43]. Perceived privacy is an independent variable affecting resident acceptance, consent quality, and long-term deployment viability.

### 5.2 Layer 1 Application

The AI model must achieve acceptable performance on ASIMOV-Agentic before any deployment consideration. Specifically, the constraint violation rate under C1 contact class scenarios (proximity and interaction with residents) must meet a defined threshold — proposed as <15% for initial provisional deployment eligibility. The ASIMOV evaluation targets the specific model version to be deployed, not a generic foundation model benchmark.

### 5.3 Layer 2 Application

An independent review body evaluates CRAB-H for the meal-round-logistics task class. An illustrative capability profile:

**PROPOSED CRAB-H ASSESSMENT — meal-round-logistics**

| Dimension | Code | Score | Note |
|---|---|---|---|
| Situational Awareness | SA | 880 | |
| Planning & Execution | UP | 840 | |
| Human Interaction | HI | 910 | |
| Self-Correction | SC | 790 | |
| Generalization | CG | 720 | ← BELOW CRAB-H FLOOR (750) |
| Energy Efficiency | CE | 850 | |
| Safety & Constraint Adherence | SCA ^H | 920 | double-weighted |
| Decision Transparency | DT ^H | 870 | double-weighted |

**Illustrative weighted profile score:** Applying CRAB-H double-weighting to SCA and DT yields a weighted mean of **857** across ten effective dimension units (Tier 1: six dimensions at 1×; Tier 2: two dimensions at 2×). Calculation: (880+840+910+790+720+850) + (920×2) + (870×2) = 8,570 ÷ 10 = 857. The profile score is reported for context only; authorization eligibility is determined by the per-dimension record, not the aggregate.

**Outcome: CONDITIONAL** — CG score of 720 falls below the proposed domain-specific floor of 750 applied in this case study illustration. Note: under CRAB-H as currently specified, mandatory minimum thresholds apply to SCA and DT (Tier 2) only. This case study illustrates how an accredited review body *could* apply domain-specific supplementary floors to Tier 1 dimensions for high-risk contact classes — a governance option left open for accreditation body discretion. A CONDITIONAL outcome on this basis would require explicit prior publication of the applicable domain floor. Earned authorization requires CG ≥ 750 at next assessment under that domain-specific floor.

The per-dimension profile communicates something a single aggregate score cannot: the system is strong on the dimensions that matter most for patient safety (SCA: 920), but its below-threshold generalization performance (CG: 720) means it should operate under enhanced supervision until deployment evidence establishes that it handles out-of-distribution situations reliably. A conditional outcome is not a rejection — it is a calibrated authorization that matches the evidence.

## 5.4 Layer 3 Application

With a conditional CRAB-H outcome, the site operator and named responsible person apply for an initial LEA authorization record:

**PROPOSED LEA RECORD — meal-round-logistics**

| Field | Value |
|---|---|
| Current | A2/C1/E1/LM1 PROVISIONAL |
| Target | A3/C1/E1/LM1 EARNED (subject to evidence review) |
| Supervised operational hours | ≥ 400 hours |
| Intervention rate | ≤ 0.25 per 100 hours |
| Fallback integrity | ≥ 98% |
| Incident condition | No unresolved S2+ incident |

The provisional record uses supervised task autonomy (A2) while site-specific evidence is accumulated. A3 is the target state after meeting the graduation thresholds, not the starting permission.

### 5.5 Governance Gap Analysis

| Layers present | Model safety verifiable? | System capability verifiable? | Operational authorization known? |
|---|---|---|---|
| ASIMOV only | Partially | No | No |
| CRAB only | No | Yes | No |
| LEA only | No | No | Partially |
| ASIMOV + CRAB | Partially | Yes | No |
| CRAB + LEA | No | Yes | Yes |
| **All three — ARC** | **✓ Yes** | **✓ Yes** | **✓ Yes** |

## 6. Discussion

### 6.1 Regulatory Alignment

ARC is relevant to the EU AI Act (Regulation (EU) 2024/1689) [9] and the EU Machinery Regulation (2023/1230) [39]. Under the AI Act, high-risk classification depends on Article 6 criteria — operation near people or somewhere in the healthcare sector does not by itself make every robotic AI system high-risk. For systems that are classified as high-risk, the Act requires risk management, data governance, technical documentation, transparency, human oversight, accuracy, robustness, cybersecurity, and post-market monitoring. The Act does not prescribe CRAB or LEA metrics; both proposals could at most become inputs to a future compliance process after independent legal and technical validation. ARC should be presented to regulators and standards bodies as a proposal for testing and comment, not a recognized compliance mechanism.

### 6.2 Trust Infrastructure

Full cryptographic verifiability of CRAB credentials requires an agreed W3C-compatible architecture: decentralized identifiers [27], verifiable credentials [28], trusted issuer registries, revocation services, and defined audit access rules. The institutional governance model — who may issue, suspend, revoke, and audit credentials — is intentionally left open for peer review and public consultation. Candidate authorities include accredited private bodies, industry or public-interest consortia, national regulators, or layered arrangements among them.

### 6.3 ROBOWELL and Empirical Calibration

LEA graduation thresholds are informed governance proposals, not empirically derived optima. The ROBOWELL programme (NordForsk, 2025–2029) — a cross-national initiative covering Norway, Denmark, and Sweden developing a Nordic Model for sustainable robot integration in residential care [45] — will generate longitudinal behavioural data from elder care deployments. Wybe Labs will initiate dialogue with ROBOWELL regarding industry collaboration. ROBOWELL deployment data, subject to ethical approval and participant consent, could provide the evidence base for calibrating LEA thresholds in the E1/C1/A2–A3 range.

### 6.4 Open Questions and Future Work

- **Integration protocol:** A formal specification of how ASIMOV outputs constrain CRAB evaluation design, and how CRAB profiles constrain LEA grant ceilings, is required.
- **Evaluation methodology:** CRAB v0.3 specifies dimensions and domain suite requirements; scenario construction, scoring, and inter-rater reliability methodology has not been fully specified — necessary for accredited body qualification.
- **Learning mode monitoring:** Audit log, behavioural consistency check, and version attestation infrastructure to enforce LEA learning mode caps requires further specification.
- **Standardization pathway:** Both CRAB and LEA are candidates for eventual IEEE or ISO standardization. LEA comment deadline: 30 November 2026. CRAB open for stakeholder input.

## 7. Conclusion

The deployment of autonomous robotic systems in safety-critical environments presents a governance challenge that existing frameworks do not adequately address. The central failure is conceptual: the conflation of capability with permission. A system's demonstrated ability to perform a task does not constitute authorization to perform that task in a specific deployment context, with a specific population, under specific regulatory constraints.

ARC — Autonomous Robotics Compliance — proposes for peer review that adequate governance requires three distinct layers:

1. **ASIMOV-Agentic** — model-level safety validation: does the AI model behave safely under physical constraints?
2. **CRAB** — system-level cognitive certification: does the integrated system meet domain-specific capability thresholds, evaluated independently?
3. **LEA** — operational-level authorization: is this specific robot authorized to perform this specific task at this specific site, on continuously reviewed evidentiary grounds?

Each layer addresses governance questions the others cannot. All three together constitute a candidate minimum viable governance architecture for safety-critical autonomous system deployment. Neither CRAB nor LEA is an adopted standard or legally recognized authorization mechanism. ARC is a starting point for peer review, pilot design, public consultation, and regulatory engagement. We invite comment on LEA-1:2026 before 30 November 2026, and welcome engagement from robotics platforms, healthcare operators, logistics deployers, and regulatory bodies.

**Capability is not permission. ARC connects both.**